\documentclass[conference]{IEEEtran}
\IEEEoverridecommandlockouts
\IEEEpubid{\makebox[\columnwidth]{978-1-6654-3902-2/21/\$31.00~\copyright2021 IEEE \hfill} \hspace{\columnsep}\makebox[\columnwidth]{ }}
\usepackage{cite}
\usepackage{amsmath,amssymb,amsfonts}
\usepackage{enumitem}
\usepackage{algorithmic}
\usepackage{graphicx}
\usepackage{textcomp}
\usepackage{xcolor}
\def\BibTeX{{\rm B\kern-.05em{\sc i\kern-.025em b}\kern-.08em
    T\kern-.1667em\lower.7ex\hbox{E}\kern-.125emX}}
    
\usepackage{geometry} 
\usepackage[utf8]{inputenc}
\usepackage[T1]{fontenc}    
\usepackage{hyperref}       
\usepackage{url}            
\usepackage{booktabs}

\usepackage{placeins}
\usepackage{adjustbox}	
\usepackage{amsmath}
\usepackage{amssymb}
\usepackage{amsthm}
\usepackage{calc}
\usepackage{wasysym}
\usepackage{babel}
\usepackage{xcolor,hyperref}
\newcommand{\hrluonote}[1]{{\color{blue}#1}}

\newcommand{\mvjnote}[1]{{\color{purple}#1}}

\usepackage{amstext}
\usepackage{mathtools}
\usepackage{graphicx}
\usepackage{xcolor}
\usepackage{booktabs}
\usepackage{array}
\usepackage{setspace}
\usepackage{bbold}
\usepackage{bm}
\usepackage{subcaption}
\usepackage{algorithm}

\theoremstyle{plain}

\numberwithin{thm}{section}
\theoremstyle{definition} 

\numberwithin{example}{section}

\numberwithin{equation}{section}

\theoremstyle{definition}

\theoremstyle{plain}

\theoremstyle{plain}

\theoremstyle{plain}

\providecommand{\corollaryname}{Corollary}
\providecommand{\definitionname}{Definition}
\providecommand{\lemmaname}{Lemma}
\providecommand{\propositionname}{Proposition}
\providecommand{\theoremname}{Theorem}

\begin{document}

\author{
    \IEEEauthorblockN{Hengrui Luo\IEEEauthorrefmark{1}, Jisu Kim\IEEEauthorrefmark{2}, Alice Patania\IEEEauthorrefmark{3}, Mikael Vejdemo-Johansson\IEEEauthorrefmark{4}}
    \IEEEauthorblockA{\IEEEauthorrefmark{1}Lawrence Berkeley National Laboratory, hrluo@lbl.gov}
    \IEEEauthorblockA{\IEEEauthorrefmark{2}DataShape team, Inria Saclay, and LMO, Universit{\'e} Paris-Saclay, jisu.kim@inria.fr}
    \IEEEauthorblockA{\IEEEauthorrefmark{3}Department of Mathematics and Statistics, University of Vermont, apatania@uvm.edu}
    \IEEEauthorblockA{\IEEEauthorrefmark{4}Department of Mathematics, CUNY College of Staten Island, \\ and Computer Science Program, CUNY Graduate Center, mvj@math.csi.cuny.edu}
}

\title{Topological Learning for Motion Data via Mixed Coordinates}

\maketitle
\IEEEpubidadjcol
\pagestyle{plain}
\begin{abstract}
{Topology can extract the structural information in a dataset efficiently. In this paper, we attempt to incorporate topological information into a multiple output Gaussian process model for transfer learning purposes. 
To achieve this goal, we extend the framework of circular coordinates into a novel framework of mixed valued coordinates to take linear trends in the time series into consideration. 

One of the major challenges to learn from multiple time series effectively via a multiple output
Gaussian process model is constructing a functional kernel. We propose
to use topologically induced clustering to construct a
cluster based kernel in a multiple output Gaussian process model.
This kernel not only incorporates the topological structural information, but also allows us to put forward a unified framework using topological information in time and motion series.
}

\end{abstract}

Periodic topological features occur naturally in modern data applications like time series \cite{tong1990non}, shapes \cite{luo_strait2021multipleshape} and motion data \cite{vejdemo2015cohomological}.  Specifically, periodic patterns can be observed in motion series data from biology, human behaviour and robotics \cite{lan2013planning,ronsse2013real,fisher1995statistical}.    

Based on current data collection practices \cite{CMU-MOCAP}, a motion series of an object can be described as a trajectory in the object’s configuration space $C$. For the human gait behavior we consider in this paper, the configuration space is a product of $SE(3)$ (the group of \emph{rigid motions} -- translations and rotations -- in $\mathbb{R}^3$, representing the position and rotation of the hip joint of a human skeleton model used for representing poses) with a collection of circle-valued angle activations for all the joints in the skeleton model. Using angles to represent the rotations in $SE(3)$, this means that a pose can be viewed as a pair $(\bm{v},\bm{\theta})$ of one position in $\mathbb{R}^3$ paired with a vector of angle values. Our main thesis in this paper is that representing such angle values as points on $S^1$ allows for a truer representation of the data without discontinuities introduced by the data representation choice itself.

\begin{figure}[ht!]
    \centering
    \includegraphics[width=.48\textwidth,keepaspectratio]{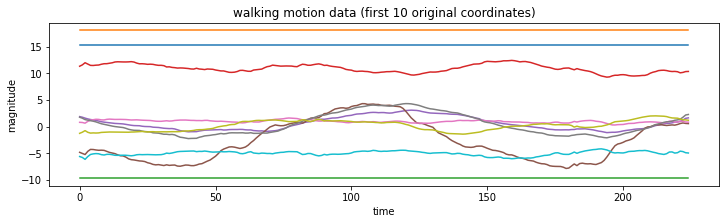}
    \caption{A sub-sampled motion data consisting of $N=10$ time series recorded by 10 sensors during a walking procedure, as provided in \cite{vejdemo2015cohomological}. We can observe that some components show more periodic characteristic and the rest components show  more linear characteristic.}
    \label{fig:motion_walking}
\end{figure}

Motions are then represented as a discrete sequence of points $(\bm{v}(t),\bm{\theta}(t))=f(t), t\in T={1,\cdots,N}$ in $C$, joining these sequence of points $f(t)$ in $C$ we yield a piecewise linear curve $f:[0,2\pi]\rightarrow C$ as a low-dimensional topological object in $C$ \cite{Luo_etal2019} shown in Figure \ref{fig:motion_walking}. In this way, even relatively simple motions can form geometrically complex
curves in the space $C$. 
Human gaits are periodic motions, since humans would repeat actions over a period of time. Such motions determine a closed curve $f$ in the configuration space and can be considered homeomorphic to a circle $S^1$ if it does not self-intersect \cite{takens1981detecting} in sufficiently high
dimensions. 

\noindent\textbf{Data Assumption.} We thus assume that our signal series can be decomposed into linear trends, periodic parts, and noise.

This type of complex motion data swamps various classical modeling and computation efforts \cite{kawahara2001periodic}. 
Existing attempts usually lack a structural understanding of the configuration space from the linear and periodic patterns exhibited in the data. 
Inspired by \cite{vejdemo2015cohomological} and previous work in statistic literature  \cite{gelfand2005bayesian,luo_strait2021multipleshape}, we plan to first generalize the circular coordinates into \emph{mixed coordinates} decomposing the linear and periodic parts of a generic series. Since the mixed valued coordinates reflect both linear and periodic patterns between series, we can describe the association and similarity of series using these coordinates. Secondly we construct a metric on the resulting coordinates for clustering series, and eventually build a transfer learning framework that uses the multiple-output Gaussian process model with a cluster-based kernel for prediction. 
\section{Mixed valued Coordinates}

As assumed above, motion time series usually arise from a periodic motion performed together with a linear space motion, possibly recorded along with noise. To encode these 
different types of signals  
in the most efficient way possible, we introduce a mixed coordinate embedding 
that separates and models these components with statistical and topological approaches.

To extract and model the linear component of the motion, we propose to use \emph{principal component analysis} (PCA) along with a correlation threshold. The classical technique of PCA summarizes the linear signals, and linear correlation thresholding allows us to select those significant linear components. Depending on different generating and signal processing mechanisms, other signal separation methods could be used (e.g., ICA \cite{hyvarinen2013independent}).

To extract and model the periodic component of the motion, we use circular coordinates. 
The circular coordinate is a topological data analysis method that constructs representative coordinate functions from cohomology classes defined on the dataset to reveal its features \cite{de2011persistent, luo_generalized_2020}. To be more precise, \emph{circular coordinates} consist of coordinate mappings with function values in $S^{1}\cong\mathbb{R}/\mathbb{Z}$, that map the dataset $X\subset \mathbb{R}^{d}$ onto a $k$-torus $\mathbb{T}^{k}=\left(S^{1}\right)^{k}$.
The circular coordinate pipeline extracts the periodic signals as circular coordinate mappings, and we pick up those significant periodic components corresponding to significantly persistent 1-cocycles. Unlike the classical correlation approach \cite{fisher1995statistical}, we can pick up more than one periodic pattern simultaneously. 

Joining these two heuristic ideas above leads us to develop the \emph{mixed coordinates} consisting of linear and circular parts, for analyzing the linear-circular association between motion series.
Formalizing the idea of the mixed coordinate method described above, we allow coordinate mappings taking values in a richer space of $\mathbb{R}^{n_\ell}\times \mathbb{T}^{n_c} $, and we call a function defined on $C$, taking values in the space of $ \mathbb{R}^{n_\ell}\times\mathbb{T}^{n_c}$ a \emph{mixed valued coordinate} associated with the dataset.

In classical time series analysis, we find mixed valued  coordinates by separating the seasonality (periodic part) and trend (linear part) of the signal in the model function \cite{brockwell2016introduction,tong1990non}. However, it has been an open problem that sinusoidal (a.k.a. trigonometric) regression would introduce Gibbs-type phenomenon and cause problems in parameter estimation \cite{eubank1990curve,quinn1989estimating,luo_strait_MOGP_2021}. Simple models in time series analysis would not capture all patterns in the motion series. In more complex models for motion series data, these issues would hinder our prediction power.

In a purely circular coordinate approach \cite{vejdemo2015cohomological}, if we discover two  significant 1-cocycles from the series; we can have two different shapes of the periodic pattern in the series (or the series can be decomposed into these two patterns represented by two 1-cocycles). However, these two different shapes do not necessarily correspond to the trend and seasonal part of the series. This kind of discrepancies create difficulties when we attempt to describe linear and circular associations in predictive modeling using topological information.
The topological approach is also fundamentally different from a statistical shape 
modeling approach to motion series \cite{wang2003automatic,luo_strait2021multipleshape}. For example, \cite{wang2003automatic} represents the motion series in a common coordinate frame and uses the Procrustes shape analysis to extract the mean shape as a signature with an intrinsic parameterization of motion. Usually, statistical shape modeling approaches do not explicitly model between-series similarities \cite{luo_strait_MOGP_2021}. 

Each motion capture time series consists of $N$ vectors of the same length $n$, where the $N$ is the number of series, while $n$ is the number of time points (or snapshots) {in each of these $N$ series}. When $N=1$, the dataset reduces to a single time series. 

We explain our approach by the following 
simple example using synthetic data. 
Suppose first that the noiseless signal of the motion time series can be written as $(f_1(t),f_2(t),f_3(t))^T=(0,-5t,\sin(7t))^T$ as a function of time $t$ with $N=3$, and add some independent white noises $\epsilon_1,\epsilon_2,\epsilon_3$ to each components and get $(f_1(t)+\epsilon_1,f_2(t)+\epsilon_2,f_3(t)+\epsilon_3)^T$.
Following our data assumption, these three coordinates are noise, linear and periodic components, respectively. 
Then we apply principal component analysis (PCA) on this $N$-dimensional data, 
yielding 3 principal components. 

Among these three principal components, we want to separate the first component (linear) from the second and third components (non-linear).
The PCA components do not inherently come with measures of their periodicity or linearity. By measuring their correlations with the time indices $t$, we can use the fact that taken over several periods, a periodic signal without a linear trend would have a much lower correlation with the time index vector than any sort of linear trend.

Here, we prefer Kendall's tau over Spearman's rho for the fact that the latter is less robust under non-normality \cite{kowalski1972effects}. 
It is natural to adopt rank correlation for separating the linear trend from the rest signals in the (ordered) time series data. We select those components with (absolute values of) correlation coefficients greater than a threshold (say, 0.50).


\begin{figure}[ht!]
    \centering
    \begin{subfigure}{0.98\linewidth}
    \includegraphics[width=\textwidth,keepaspectratio]{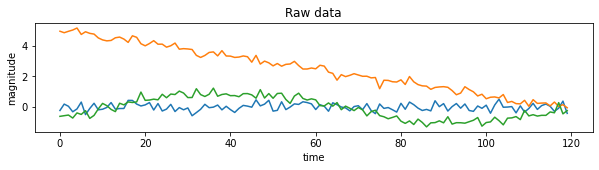}
    \caption{}
    \end{subfigure}
    \begin{subfigure}{0.98\linewidth}
    \includegraphics[width=\textwidth,keepaspectratio]{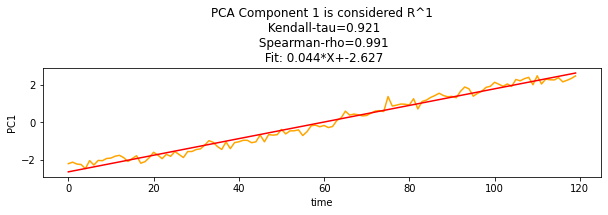}
    \caption{}
    \label{subfig:PCA_noisy_first}
    \end{subfigure}
    \begin{subfigure}{0.98\linewidth}
    \includegraphics[width=\textwidth,keepaspectratio]{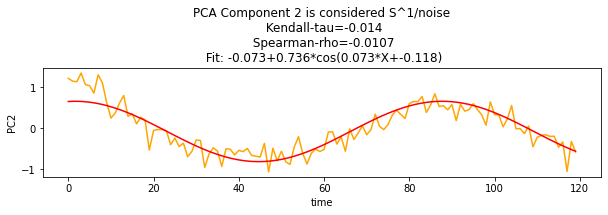}
    \caption{}
    \label{subfig:PCA_noisy_second}
    \end{subfigure}
    \begin{subfigure}{0.98\linewidth}
    \includegraphics[width=\textwidth,keepaspectratio]{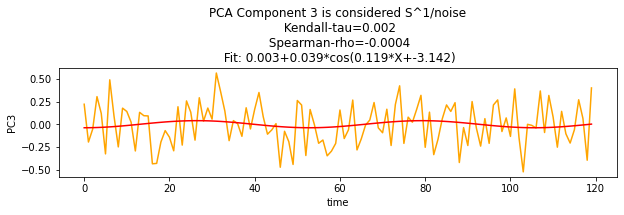}
    \caption{}
    \label{subfig:PCA_noisy_third}
    \end{subfigure}
    \caption{The principal components (ordered by magnitude of loading factors) for the motion data $(\epsilon_1,-5t+\epsilon_2,\sin(7t)+\epsilon_3)$, where $\epsilon_1,\epsilon_2,\epsilon_3$ are normal noise with 0 mean and standard deviation 0.2.\\
    The 1st component in (\subref{subfig:PCA_noisy_first}) is considered linear while the 2nd and the 3rd components in (\subref{subfig:PCA_noisy_second}) and (\subref{subfig:PCA_noisy_third}) are considered periodic or noise. \\ The 1st and 2nd components are quite strong, while the 3rd component seems noisy. \\
    The x-axis is for indices $t$; the y-axis is for series value. The orange lines show the series points joined by segments. The red solid lines are rough sketches of linear or trigonometric regression fit.}
    \label{fig:PCA_on_noisy_signal}
\end{figure}

However, we still need to separate the periodic part from the possible noise showing up as the third component in Figure \ref{fig:PCA_on_noisy_signal}. Instead of resorting to statistics, we use circular coordinates which are better suited to capture periodic feature from background noise. This also follows a thresholding procedure, but with topological persistence. 

In particular, for each remaining components, we first use the canonical  $1-$(or $r-$)delay embedding \cite{tong1990non,ragwitz2002markov} for a series and put it into a point set in $\mathbb{R}^2$ (or $\mathbb{R}^{r+1}$) in the form of $(X_t, X_{t+\varepsilon}, ..., X_{t+r\varepsilon})$ ($\varepsilon=1$ by default). Then we can perform the circular coordinate computation for this resulting point set \cite{vejdemo2015cohomological,luo_generalized_2020}. By looking at the persistence of each cohomological feature 
(keeping features with persistent lifetimes longer than a prescribed threshold),
we can differentiate the periodic signals (with long persistence) from noise (with short persistence). 

\noindent\textbf{Computation of Mixed Valued Coordinates.}
\begin{enumerate}[leftmargin=*,label={(S\arabic*)}]
    \item (De-trend, optional) For a motion series $(f_1(t),\cdots,f_N(t))^T,f_i(t)\in\mathbb R^{1\times n},t=t_1,\cdots,t_n$ collected from $N$ different series, we fit each component $f_i$ to a simple linear regression $\hat{f_i}$ with predictor $t$ and calculate the residual vector $(f_1(t)-\hat{f}_1(t),\cdots,f_N(t)-\hat{f}_N(t))^T=(r_1(t),\cdots,r_N(t))^T$.
    \item (Separation) Perform PCA on the residual vector $(r_1(t),\cdots,r_N(t))^T$ and obtain the principal components $(pr_1(t),\cdots,pr_{N'}(t))^T,N'\leq N$ (Usually we take $N'=N$, and since $N'$ represents the retained coordinate functions, $N' = n_\ell+n_c+n_n$ where:
    \begin{enumerate}
    \item $n_\ell$ is the number of linear components; 
    \item $n_c$ is the number of periodic components;
    \item $n_n$ is the number of components representing pure noise.
  \end{enumerate}
    The (absolute values of) Kendall's $\tau$ coefficients of  $\tau(t,pr_i(t)),i=1,\cdots,N'$ are used to determine $n_\ell$ linear  components to retain. \\
    To be precise, we consider the components with large Kendall coefficients as linear parts (as vectors  $\bar{pr}_1,\cdots,\bar{pr}_{n_\ell} $) ; while the $(n_c+n_n)=N'-n_\ell$ components with small Kendall coefficients as periodic parts and potential noise (as vectors  $\bar{pr}_{n_\ell+1},\cdots,\bar{pr}_{N'}$). 
    Note that $\{\bar{pr}_i\}_{i\leq N'}$ is a renumbering of $\{pr_i\}_{i\leq N'}$ so that the linear principal components come first.
    \item (Juxtaposition) 
    The first part of the mixed coordinate is 
    $\bar{pr}_1,\cdots,\bar{pr}_{n_\ell}$, which are the linear principal components obtained from S2.\\
    The second part of the mixed coordinate is 
    $c_{n_\ell+1},\cdots,c_{n_\ell+n_c}$, obtained by dropping the noise components in the circular coordinates pipelines for $\bar{pr}_{\ell+1},\cdots,\bar{pr}_{N'}$ and drop the principal components based on the  persistence of 1-cocycles for this component.\\
    \hspace{4mm}Precisely, we first compute a delay embedding of each principal component series $\bar{pr}_{\ell+1},\cdots,\bar{pr}_{N'}$, and then compute circular coordinates for each.
    If there are significant 1-cocycles with persistence greater than a prescribed threshold, then the corresponding principal component is considered as periodic and the circular coordinates are kept; otherwise the corresponding principal component is considered to be a noisy component and dropped. 
\end{enumerate}
\begin{figure}
\begin{center}

\noindent\fbox{\begin{minipage}[t]{1\columnwidth - 2\fboxsep - 2\fboxrule}%
\begin{center}{\footnotesize{}}%
\fbox{\begin{minipage}[t]{0.95\columnwidth}%
{\footnotesize{}$\text{(S1)}$}{\footnotesize\par}

{\footnotesize{}
\begin{align*}
\ensuremath{(f_{1}(t),\cdots,f_{N}(t))^{T}} & ,f_{i}(t)\in\mathbb{R}^{1\times n},t=t_{1},\cdots,t_{n}\\
\Downarrow & \text{de-trend (optional)}
\end{align*}
}{\footnotesize\par}%
\end{minipage}}{\footnotesize\par}

{\footnotesize{}}%
\fbox{\begin{minipage}[t]{0.95\columnwidth}%
{\footnotesize{}$\text{(S2)}$}{\footnotesize\par}

{\footnotesize{}
\begin{align*}
\ensuremath{(r_{1}(t),\cdots,r_{N}(t))^{T}} & =(f_{1}(t)-\hat{f}_{1}(t),\cdots\\
 & ,f_{N}(t)-\hat{f}_{N}(t))^{T}\\
\Downarrow & \text{PCA}\\
\ensuremath{(pr_{1}(t),\cdots,pr_{N}(t))^{T}} & ,\\
\Downarrow & \text{possibly dropping some PCs}\\
\ensuremath{(pr_{1}(t),\cdots,pr_{N'}(t))^{T}} & ,N'\leq N\\
\Downarrow & \text{ Kendall's tau (and re-arrange)}\\
\ensuremath{(\bar{pr}{}_{1}(t),\cdots,\bar{pr}{}_{n_\ell}(t)} & \ensuremath{\quad\quad\bar{pr}{}_{n_\ell+1}(t),\cdots,\bar{pr}{}_{N'}(t))^{T}}\\
\text{linear component}\Downarrow & \quad\quad\Downarrow\text{periodic+\ensuremath{\epsilon} component}
\end{align*}
}{\footnotesize\par}%
\end{minipage}}{\footnotesize\par}

{\footnotesize{}}%
\fbox{\begin{minipage}[t]{0.95\columnwidth}%
{\footnotesize{}$\text{(S3)}$
\begin{align*}
\Downarrow & \quad\quad\Downarrow\text{circular coordinates}\\
 & \quad\quad c{}_{n_\ell+1}(t),\cdots,c{}_{N'}(t)\\
\Downarrow & \quad\quad\Downarrow\text{drop components }\\
 & \quad\quad\quad\text{ with low persistence}\\
\ensuremath{(\bar{pr}{}_{1}(t),\cdots,\bar{pr}{}_{n_\ell}(t)} & \quad\quad c{}_{n_\ell+1}(t),\cdots,c{}_{n_\ell+n_c}(t))^{T}
\end{align*}
}%
\end{minipage}}{\footnotesize\par}

\end{center}%
\end{minipage}}

\end{center}

\caption{Procedural chart for the computation of mixed coordinates.}
\end{figure}

In this example in Figure \ref{fig:PCA_on_noisy_signal}, we do not perform the optional de-trending, but for intertwined signals like $-2t+\sin(4t)$, detrending may be needed. The number of total components is $N=3$, and we retain $N'=3$ principal components in step (S1). Among these $N'=3$ principal components, we have $n_\ell=1$ linear component (i.e., principal component 1 in Figure \ref{fig:PCA_on_noisy_signal}(\subref{subfig:PCA_noisy_first})) separated from $n_c+n_n=2$ periodic or noisy components (i.e., principal components 2,3 in Figure \ref{fig:PCA_on_noisy_signal}(\subref{subfig:PCA_noisy_second}) and (\subref{subfig:PCA_noisy_third})) in step (S2). In these $n_c+n_n=2$ periodic and noise components; we separate $n_c=1$ periodic component (i.e., Figure \ref{fig:PCA_on_noisy_signal}(\subref{subfig:PCA_noisy_second}) ); and take the last component as a noisy component (i.e., Figure \ref{fig:PCA_on_noisy_signal}(\subref{subfig:PCA_noisy_third}) ) using the threshold test on its circular coordinates as described in (S3).

The final mixed coordinates \\
\resizebox{1.0\columnwidth}{!}{
$
(\bar{pr}{}_{1}(t),\cdots,\bar{pr}{}_{n_\ell}(t),c{}_{n_\ell+1}(t),\cdots,c{}_{n_\ell+n_c}(t))^{T}\in\mathbb{R}^{n_\ell}\times\mathbb{T}^{n_c}
$
} \\
consist of two parts.

The first part $\bar{pr}{}_{1}(t),\cdots,\bar{pr}{}_{n_\ell}(t)$ is extracted from the motion series consisting of $(f_1,\cdots,f_N)$. These are the linear parts of the motion series, determined by Kendall's tau.
The second part $c{}_{n_\ell+1}(t),\cdots,c{}_{n_\ell+n_c}(t)$ is ``significantly persistent'' circular coordinates that separate periodic patterns from noise. 

In the simplest case {where $n_\ell+n_c=N,n_n=0$,
{(i.e., we do not assume noise, but keep all principal components, take $n_\ell$ components as linear trends, and the remaining $n_c$ components as significant periodic part.)} 
the PCA rotates the $N\times n$ motion series dataset into such a position that for each time point the variations of the series corresponding to the leading principal component are maximized. 
A linear trend will produce a monotonic sequence of values, while a seasonality component will produce a more ``flat'' sequence of values, a difference that can be measured well using Kendall's tau coefficients.
By looking at the persistence of circular coordinates, we separate the periodic signals from the noise. 

\section{Metrics of Topological Coordinates\label{sec:metrics of topological coords}}

After computing the mixed coordinates, we need a metric between coordinates to reflect the topological resemblance of motion datasets. 
The need for a metric is not only motivated by classification as mentioned in \cite{vejdemo2015cohomological}, but also inspired by the cluster regularization method for kernel construction in the Gaussian Processes literature \cite{jacob2008_clustering}. 

The mixed coordinates consist of a linear part and circular coordinates, so we need a metric that reflects both. This can be done by considering a product of metrics $\mathcal D = d'\times d$, where $d'$ is for the linear part and $d$ is for the circular coordinates. Since we can use typical metrics such as $L_{2}$ to compute distances for the linear part $d'$, in this section we will focus on constructing a metric $d$ for circular coordinates.

{
Circular coordinates can be considered as representatives of equivalence classes \cite{luo_generalized_2020} and a metric needs to take this into consideration -- most commonly this is done by minimizing over representative cocycles of the equivalence classes, or by enforcing properties that are invariant over the equivalence classes. Our goal is for two circular coordinates computed from two datasets that are topologically similar to have a smaller distance in such a metric than for a pair of topologically dissimilar datasets. 
}

We consider two circular coordinates that are from different motion series of possibly different length and let us assume for convenience that $n_{c}=1$, and that we can treat each circular coordinates value $c(t_{i})\in \mathbb{S}^{1}$ as  embedded in $\mathbb{R}$.

Let $t=(t_{1},\ldots,t_{n}),t'=(t'_{1},\ldots,t'_{m})$ be two sequences of time series indices, and let  
$c(t),c'(t)\in\mathbb{R}^{n},
\tilde{c}(t'),\tilde{c}'(t')\in\mathbb{R}^{m}$ be circular coordinates 
of possibly different lengths.
There are several desired properties that a metric $d$ between two
circular coordinates should satisfy:  
\begin{enumerate}[label={(P\arabic*)}]
\item Mod 1 invariance, that is, the distance is invariant under Mod 1: i.e., for circular coordinates $c(t),c'(t)\in \mathbb{R}^{n},t=t_1,\cdots,t_n$ and $\tilde{c}(t'),\tilde{c}'(t')\in \mathbb{R}^{m},t'=t'_1,\cdots,t'_m$, if $c=c'\pmod1$ and $\tilde{c}=\tilde{c}'\pmod1$ then $d(c,\tilde{c})=d(c',\tilde{c}')$. This is because the value of circular parts are all supposed to be on $S^1$.
\item Inversion invariance, if $c(t_{i+1})-c(t_{i})=-(c'(t_{i+1})-c'(t_{i}))$
for all $i$, then $d(c,c')=0$. 
\item Translation invariance, 
for any $a,b\in\mathbb{R}$, $d(c+a,\tilde{c}+b)=d(c,\tilde{c})$. 
\end{enumerate}

To ensure that the above properties (P1)-(P3) hold, we define several transforms that help to properly reflect the topological resemblance between circular coordinates.

For (P1), we define a transform on circular coordinates as shifting the circular coordinate values so that the neighboring values are always close enough, i.e., to satisfy that $-0.5<c_{i+1}-c_{i}\leq0.5$. For this, we define the transform $T_{M}$ as: 
\begin{enumerate}[label=\roman*]
\item $(T_{M}(c))(t_0)=c(t_0)$ 
\item $(T_{M}(c))(t_{i+1})=c(t_{i+1})+k$ for $k\in\mathbb{Z}$ that satisfies $-0.5<(T_{M}(c))(t_{i+1})-(T_{M}(c))(t_{i})\leq0.5$.
\end{enumerate}
For (P2), we define a transform on circular coordinates as inverting the circular values so that $c(t_{n-1})-c(t_{0})$ is
always positive. For this, we define the transform $T_{I}$ as: 
\begin{enumerate}[label=\roman*]
\item if $c(t_{n-1})>c(t_{0})$, then $T_{I}(c)=c$. 
\item if $c(t_{n-1})<c(t_{0})$, then $(T_{I}(c))(t_{0})=c(t_{0})$
and $(T_{I}(c))(t_{i+1})=(T_{I}(c))(t_{i})+c(t_{i})-c(t_{i+1})$. 
\end{enumerate}
Figure~\ref{fig:metric_transform} demonstrates an example how applying $T_{M}$ and $T_{I}$ helps the transformed circular coordinates to be inversion invariant. 

For (P3), 
given a metric between two vectors, we define a transform on the metric as follows: the transformed metric compares the circular coordinates $c$ to the circular coordinates $\tilde{c}+a$ for $a$
varying from $\min{c}-\min{\tilde{c}}$ to $\max{c}-\max{\tilde{c}}$. In other words,
it is to add the offset before comparing two circular
coordinates and return the minimum possible distance. Precisely, we
define a transform $T_{L}$ as: 
\[
T_{L}(d)(c,\tilde{c})= 
\min_{\min{c}-\min{\tilde{c}}\leq a \leq \max{c}-\max{\tilde{c}}}d(c,\tilde{c}+a).
\]
This makes it a 1-dimensional optimization problem: just find the offset that minimizes the signal difference. 

Combining $T_{M}$, $T_{I}$ and $T_{L}$ into $\Phi$ as 
\[
\Phi(d)(c,\tilde{c}):=T_{L}(d)(T_{I}\circ T_{M}(c),T_{I}\circ T_{M}(\tilde{c}))
\]
We obtain a transform that satisfies (P1), (P2), and (P3). In other words,
$c=c'\pmod1$ and $\tilde{c}=\tilde{c}'\pmod1$ then $\Phi(d)(c,\tilde{c})=\Phi(d)(c',\tilde{c}')$,
if $c(t_{i+1})-c(t_{i})=-(c'(t_{i+1})-c'(t_{i}))$ for all $i$ then $\Phi(d)(c,c')=0$, and for any $a,b\in\mathbb{R}$,
$\Phi(d)(c+a,\tilde{c}+b)=\Phi(d)(c,\tilde{c})$.  

Then, as discussed in the beginning, we define the metric $\mathcal D = d'\times d$, where $d'$ is a chosen metric for the linear part and $d$ is the topological metric with the desired properties we discussed in this section. 

\begin{figure}
    \centering
    \begin{subfigure}{0.48\linewidth}
    \centering
    \includegraphics[width=\linewidth,keepaspectratio]{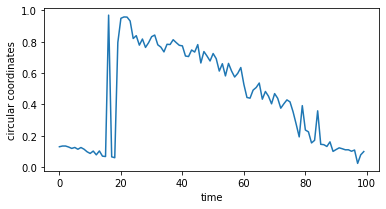}
    \caption{Circular coordinates.}
    \label{fig:metric_transform_1_before}
    \end{subfigure}
    \hfill
    \begin{subfigure}{0.48\linewidth}
    \centering
    \includegraphics[width=\linewidth,keepaspectratio]{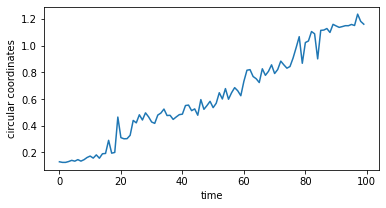}
    \caption{$T_{I} \circ T_{M}$ applied to (\subref{fig:metric_transform_1_before}).}
    \label{fig:metric_transform_1_after}
    \end{subfigure}
    \begin{subfigure}{0.48\linewidth}
    \centering
    \includegraphics[width=\linewidth,keepaspectratio]{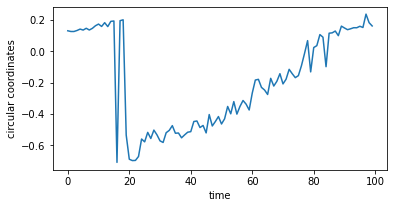}
    \caption{Depicts (\subref{fig:metric_transform_1_before}) inverted.}
    \label{fig:metric_transform_2_before}
    \end{subfigure}
    \hfill
    \begin{subfigure}{0.48\linewidth}
    \centering
    \includegraphics[width=\linewidth,keepaspectratio]{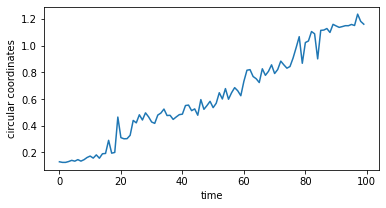}
    \caption{$T_{I}\circ T_{M}$ applied to (\subref{fig:metric_transform_2_before}).}
    \label{fig:metric_transform_2_after}
    \end{subfigure}
\caption{ (\subref{fig:metric_transform_1_before}) is the circular coordinates of the motion data in Figure \ref{fig:PCA_on_noisy_signal} from $20$-delay embeddings, and  (\subref{fig:metric_transform_2_before}) is its inversion. (\subref{fig:metric_transform_1_after}) and (\subref{fig:metric_transform_2_after}) correspond to the transform  
$T_{I}\circ T_{M}$ applied to (\subref{fig:metric_transform_1_before}) and (\subref{fig:metric_transform_2_before}). The graphs (\subref{fig:metric_transform_1_before}) and (\subref{fig:metric_transform_2_before}) are quite different, but (\subref{fig:metric_transform_1_after}) and (\subref{fig:metric_transform_2_after}) are similar. 
This illustrates the inversion invariance induced by the transform $T_I\circ T_M$.
}
\label{fig:metric_transform}
\end{figure}

\section{Multiple-output Gaussian Process Modeling}


The clustering based on a mixed coordinate can clearly be used for classification along with the metric above. However, in this section, we propose that the clustering information also helps in building a generative prediction model.

Multiple-output Gaussian process (GP) modeling is a fundamental statistical model used in transfer learning \cite{rasmussen2003gaussian,sid2019multitask,GPTuneUserGuide}. For multiple series as 
functional or curve data
, we use the following model for motion series at time $t$,
\begin{align} 
(f_1(t),f_2(t),\cdots,f_N(t))^T+\bm{\epsilon},f(t)\in C,
\label{eq:moGP}
\end{align}
with $\bm{\epsilon}$ a random vector of added noise modeled by a Gaussian Process.
This notation for a vector-valued model can be understood as a Gaussian Process in two ways: either as an ensemble of single-output Gaussian processes, or as a single multiple-output Gaussian Process. The distinction is most visible in the covariance kernel for the $\bm{\epsilon}$: in the single-output ensemble case, the individual Gaussian Processes would be assumed to be decorrelated, making the covariance kernel represented by a diagonal matrix, while the multiple-output Gaussian Process allows us to model correlations between the $f_1,\dots,f_N$ in the covariance kernel which enhances the estimation (for denoising) and prediction abilities of the resulting model. \cite{rasmussen2003gaussian}

There are various possible choices of covariance kernels for the multiple-output kernels \cite{alvarez2011kernels}. For our work, we choose the cluster-based kernels \cite{jacob2008_clustering}, and plan to use a cluster-based regularized kernel that clusters using our metric of mixed coordinates.
This way, we can group similar curves together by their shape descriptors, that is, metrics for mixed coordinates.
More formally, we split $M$ curves into $r$ different clusters, and specify a regularizer which limits the complexity of the fitted multivariate GPs $\bm{{\displaystyle f}}=(f_{1},\cdots,f_{M})$,  so that the covariance matrix would be block-diagonal and the mean curve $\bar{f}_{c}$ would be a good representative of the group. 

For the cluster-based regularizer applied to multiple-output GP models on $\bm{f}$, the regularization term
$R(\bm{f})$ that reflects the "in-cluster" and "between-cluster" trade-off could be written as : 
\begin{align}
\displaystyle R(\bm{f})=\lambda_{1}\sum_{c=1}^{r}\sum_{l\in I(c)}\mathcal{D}(f_{l},\bar{f_{c}})+\lambda_{2}\sum\limits _{c=1}^{r}m_{c}\mathcal{N}(\bar{f_{c}}),
\end{align}
where $\lambda_{1},\lambda_{2}>0$ are tunable weighting parameters,
\begin{itemize}
\item ${\displaystyle I(c)\subset\{1,\ldots,M\}}$ is the index set of principal components that belong to cluster ${\displaystyle c}$. The collection of sets $I(c)$ forms a partition of $\{1,\dots,M\}$. 
\item ${\displaystyle m_{c}}$ is the cardinality of cluster $c$, and by
the decomposition above, $\sum_{c=1}^{r}m_{c}=M$. 
\item ${\displaystyle \bar{f_{c}}=\frac{1}{m_{c}}\sum\limits _{q\in I(c)}f_{q}}$
is the average centroid time series of the cluster $c$.
\end{itemize}
This regularizer divides the components into $r$ clusters and forces the components in each cluster of (closed) curves to be similar.

As a next step, we plan to explore the use of the mixed coordinate induced metric $\mathcal D = d\times d'$ and the corresponding norm $\mathcal N$  for regularization, in the case where multiple registrations are necessary among a collection of series. 

This echoes the practice of studying linear-circular associations in \cite{fisher1995statistical}. The classical literature restricts to learning only circular-circular, linear-linear or circular-linear associations separately, in the current modeling, we use the mixed coordinate and the metric between coordinates to study all three kinds of associations simultaneously.


\section{Conclusion}
In this paper, we propose an extension from circular coordinates to mixed coordinates that incorporates topological and non-topological information in models for motion capture data.
We regularize a Gaussian Process kernel using a topologically induced clustering approach.
Although we have obtained preliminary results showing successes, more experiments and refined data analysis are needed for verifying the performance as a transfer learning model. 
We further expect to develop an asymptotic theory and consistency results for this Gaussian Process model.

Motivated by the data representation in motion capture datasets,
we first generalize the circular coordinates into mixed coordinates that take values in $ \mathbb{R}^{n_\ell}\times\mathbb{T}^{n_c}$. 
We introduce a metric adapted to work with these mixed coordinates and use this metric to cluster the motion capture time series.
Finally, we leverage the clustering information to construct a covariance kernel in a multiple-output GP model, showing the power of transfer learning when the topology of the data is taken into account. 

This paper joins the current trend to develop methods that synthesize both topological and geometrical information.
In future works, we plan to combine the ideas of this paper with our recent extension of the circular coordinate paradigm to generalized circular coordinates (GCC) \cite{luo_generalized_2020}.
It is also interesting to compare other metrics, like  Gromov-Wasserstein \cite{chowdhury2021generalized}, for better construction of the covariance kernel. 
A challenge we have started to approach is how to derive suitable metrics for data when there are different aspects concurrently present -- like the circular and linear coordinates in this work.

\section*{Acknowledgement}
HL was supported by the Director, Office of Science, of the U.S. Department of Energy under Contract No. DE-AC02-05CH11231.

Our code for mixed coordinate implementations and experiments is publicly available at \url{https://github.com/hrluo/TopologicalMotionSeries}.

\clearpage
\bibliographystyle{ieeetr}
\bibliography{mixedCoordinates}

\end{document}